\documentclass[manuscript,screen]{acmart}
\AtBeginDocument{%
  }

\setcopyright{acmlicensed}
\copyrightyear{2026}
\acmYear{2026}
\acmDOI{XXXXXXX.XXXXXXX}
\acmISBN{978-1-4503-XXXX-X/2018/06}

\renewcommand\footnotetextcopyrightpermission[1]{} 

\makeatletter
\let\@authorsaddresses\@empty
\makeatother

\begin{document}

\title{Mitigating Bias in Concept Bottleneck Models for Fair and Interpretable Image Classification}

\author{Schrasing Tong}
\email{st9@mit.edu}
\affiliation{%
 \institution{Massachusetts Institute of Technology}
 \city{Cambridge}
 \state{Massachusetts}
 \country{United States}}

\author{Antoine Salaun}
\affiliation{%
 \institution{Massachusetts Institute of Technology}
 \city{Cambridge}
 \state{Massachusetts}
 \country{United States}}

\author{Vincent Yuan}
\affiliation{%
 \institution{Massachusetts Institute of Technology}
 \city{Cambridge}
 \state{Massachusetts}
 \country{United States}}

\author{Annabel Adeyeri}
\affiliation{%
 \institution{Massachusetts Institute of Technology}
 \city{Cambridge}
 \state{Massachusetts}
 \country{United States}}

\author{Lalana Kagal}
\affiliation{%
 \institution{Massachusetts Institute of Technology}
 \city{Cambridge}
 \state{Massachusetts}
 \country{United States}}

\renewcommand{\shortauthors}{Tong et al.}

\begin{abstract}
Ensuring fairness in image classification prevents models from perpetuating and amplifying bias.  
Concept bottleneck models (CBMs) map images to high-level, human-interpretable concepts before making predictions via a sparse, one-layer classifier. 
This structure enhances interpretability and, in theory, supports fairness by masking sensitive attribute proxies such as facial features.
However, CBM concepts have been known to leak information unrelated to concept semantics and early results reveal only marginal reductions in gender bias on datasets like ImSitu.  
We propose three bias mitigation techniques to improve fairness in CBMs: 1. Decreasing information leakage using a top-k concept filter, 2. Removing biased concepts, and 3. Adversarial debiasing.
Our results outperform prior work in terms of fairness-performance tradeoffs, indicating that our debiased CBM provides a significant step towards fair and interpretable image classification.
\end{abstract}



\maketitle

\section{Introduction}
\label{sec:intro}

In recent years, computer vision systems have achieved remarkable advancements, making breakthroughs in different application domains such as healthcare, autonomous vehicles, and social media. 
While these technologies increase efficiency and bring societal benefits, growing evidence highlights significant fairness concerns that result from models perpetuating and amplifying existing bias. 
Studies showing biased facial recognition~\cite{buolamwini2018gender,balakrishnan2021towards} and unequal performance across demographic groups~\cite{seyyed2020chexclusion,sudhakar2023icon} raise critical questions about the ethical implications and impact of these systems. 

To address these issues, researchers have devised various methods for bias mitigation, often building off techniques in the fairness subfield. 
For example, \cite{joo2020gender,bendekgey2021scalable} focus on creating a more balanced training dataset and \cite{wang2019balanced} leverages adversarial debiasing to remove the influence of sensitive attributes. 
Since most state-of-the-art methods in this space rely on deep neural networks, interpretability becomes crucial to understanding model behavior and the improvements made after bias mitigation techniques~\cite{tian2022image}. 
Intuitively, for most image classification tasks unrelated to facial recognition, too much detail on the person in the scene is often unnecessary and susceptible to bias. 
The model should make predictions based on discriminating features for the task and the person's actions rather than facial features or in some cases sensitive attribute proxies like clothing. 

To this end, Concept Bottleneck Models (CBMs)~\cite{koh2020concept} first learn a mapping from input images to a layer of human specified concepts before making predictions using a sparse, one layer network on these interpretable concepts. 
Doing so effectively masks unnecessary details while achieving interpretability through concept weights, similar to linear regression. 
Despite its benefits, CBMs are known to have information leakage~\cite{mahinpei2021promises} in which the concepts encode hidden information unrelated to concept semantics; this includes information on sensitive attributes as well. 
To decrease human effort in generating and creating ground truth labels for concepts, we use a more efficient version of the Label-free CBM framework~\cite{oikarinen2023label}, leveraging GPT-3 and Contrastive Language-Image Pre-Training (CLIP)~\cite{radford2021learning}.
We evaluate fairness-performance tradeoffs on the ImSitu Dataset~\cite{yatskar2016} for action recognition to simulate real-world scenarios and take advantage of gender ground truths. 
We adopt data and model leakage metrics from \cite{wang2019balanced} to quantify fairness by focusing on preventing bias amplification.
We compare fairness-performance tradeoffs for three models: zero-shot learning with CLIP, neural network trained on embeddings from CLIP's image encoder, and CBM using CLIP's inference on concepts. 
Results show that CBMs improve both fairness and interpretability compared to an optimized DNN while slightly reducing accuracy. 
However, information leakage creates a tradeoff between fairness, interpretability, and performance: The model requires a certain number of concepts to achieve strong performance but doing so leaks information on sensitive attributes and makes interpretation more difficult. 

The main contribution of this paper is to develop bias mitigation 
algorithms for CBMs that improve fairness and interpretability. 
More specifically, we introduce three techniques in this paper: 1. Decreasing information leakage and further increasing interpretability through sparsity or a top-k filter for concept activations. This approach has the additional benefit of not relying on sensitive attribute ground truths during debiasing since labeling attributes can be expensive and prone to human bias in practice. 2. Removing concepts that are more closely correlated with sensitive attributes, and 3. Applying adversarial debiasing to CBMs to optimize for fairness performance tradeoffs. 
The top-k concept filter we proposed outperforms label-free CBM's sparsity on the FC layer in terms of fairness-performance tradeoffs but still does not eliminate information leakage completely. 
Using a gender classifier to identify and remove biased concepts or asking a LLM to self-rate biased concepts based on semantics do not produce strong results. 
Furthermore, the model learns to leak gender information through different concepts during re-training, thus concept removals have to be implemented by zeroing out concepts at test time. 
Adversarial debiasing optimizes for fairness-performance tradeoffs and results outperform that of prior work. 
Compared to black-box models, the shift in concept weights and contributions provides transparency into model behavior and the debiasing process. 
In general, the top-k concept filter achieves strong fairness-performance tradeoffs without the need for sensitive attribute ground truths during debiasing and applying adversarial debiasing improves results further, decreasing bias by 28\% with very minor accuracy loss.

\section{Related Work}

\subsection{Fairness in Image Classification}
Bias in image datasets can take many different forms, such as image quality or location taken, and have been known to negatively affect the generalization performance of trained models~\cite{torralba2011unbiased,tommasi2017deeper}.
Fairness centers on the subset of bias surrounding humans and sensitive attributes, for example associating professions with gender or recognizing objects poorly for other cultures. 
Such issues can produce serious consequences when the algorithm achieves different levels of performance or even discriminates against people~\cite{hendricks2018women,buolamwini2018gender,ryu2017improving}.
Researchers have studied fairness in a variety of computer vision tasks, including facial recognition~\cite{buolamwini2018gender,georgopoulos2021mitigating}, object detection~\cite{sudhakar2023icon}, medical imaging~\cite{seyyed2020chexclusion}, and representation learning~\cite{park2023training,zhang2023learning}.
In general, the presence of multiple sensitive attributes represented in high dimensional domains makes both detecting bias in trained models~\cite{li2021discover,hu2020crowdsourcing} and collecting more balanced image datasets~\cite{xu2021consistent,ramaswamy2021fair} hard. 

Prior work on improving fairness can be categorized into several approaches. 
Incorporating an additional fairness loss constraint can optimize for the corresponding fairness metric~\cite{chang2020adversarial,wang2019balanced,chuang2021fair}.
The most notable example leverages adversarial debiasing to eliminate the sensitive attribute cues in the model during training~\cite{wang2019balanced}.
Another approach focuses on creating a more balanced training dataset with equal representations.
For example, generating synthetic images by manipulating latent features~\cite{joo2020gender}, optimizing the sample selection procedure~\cite{bendekgey2021scalable}, or reweighing samples~\cite{zhao2020maintaining}.
Last but not least, researchers have attempted to learn independent features for the given task that do not contain information on the sensitive attributes~\cite{liu2018exploring,rifai2012disentangling,liu2017adaptive}. 
This approach often uses representation learning techniques to minimize the reconstruction loss as well as the similarity between the learned features and the sensitive attributes.

Despite recent advancements, there are several limitations with current approaches to fairness. 
Representation learning raises concerns over utility~\cite{moyer2018invariant} and applicability in practice~\cite{caton2024fairness}; most research focuses on facial analysis, for example on the CelebA Dataset~\cite{liu2018large}, where well-aligned features exist. 
For general real-world applications, \cite{wang2019balanced,xu2021consistent} have observed that trained models can still produce unfair predictions despite learning from balanced datasets. 
Detecting the proxy features behind this issue can help us gain deeper understanding of the source of the bias and devise more interpretable methods for bias mitigation that go beyond optimizing for a single loss metric.  
To this end, CBMs provide a novel approach that prioritizes interpretability yet takes insight from representation learning by hiding overly detailed information that may serve as sensitive attribute proxies. 

\subsection{Concept Bottleneck Models}
Interpretability allows us to understand more about the decision process of a deep learning model, in this case where the bias occurred. 
Since interpretable methods and explanations only approximate model behavior, saliency maps~\cite{selvaraju2017grad} and TCAV~\cite{kim2017interpretability} have been shown to provide valuable but sometimes inaccurate insights to fairness~\cite{tong2020investigating}.
CBMs incorporate interpretability by first predicting human understandable concepts and then using those concepts to make the final prediction. 
For example, when predicting bird species, a biology expert can come up with concepts on color, beak shape, feet, and body shape to make predictions based on domain knowledge. 
This method combines the advantages of explanations with human specified concepts~\cite{bau2017network,kim2017interpretability,ghorbani2019towards} and a more interpretable final layer~\cite{wong2021leveraging}. 
Since interpretability is built in by design, CBMs improve upon post-hoc explanation methods~\cite{ribeiro2016should,selvaraju2017grad,simonyan2013deep} by removing the error introduced when approximating the model's behaviors. 
Last but not least, CBMs support interventions on model predictions that are based on concepts, relating to prior work on model editing~\cite{bau2020rewriting}.
We believe that these features can help us achieve both fairness and interpretability in image classification.
CBMs remain an active field of research and we chose label-free CBM over newer models~\cite{srivastava2024vlg} for its wider applicability in practice.  

\section{Methodology}

\subsection{Dataset and Pre-processing}
To mimic fairness issues in image classification in a real-world setting, we evaluate the CBMs using the ImSitu Dataset. 
ImSitu is one of the only real-world, single label image classification datasets with ground truths for sensitive attributes, which allows us to measure fairness and compare with prior research. 
Each image is labeled with 1. the main activity (verb), 2. the participating actors, objects, substances, and locations, and 3. the roles these participants play in the action. 

Since fairness focuses on human subjects, we filtered out images that have non-human Agent metadata. 
More specifically, we parsed images with Agent containing any of the following words as male: ['man', 'male', 'boy', 'mister', 'father', 'brother', 'uncle', 'husband', 'son', 'dad', 'groom'], and similarly ['woman', 'female', 'girl', 'miss', 'mom', 'sister', 'mother', 'aunt', 'wife', 'daughter', 'bride'] as female. 
Due to class imbalance, we kept the 200 most represented verbs to ensure sufficient sample sizes. 
The pre-processed dataset has 20792 images, with 10794 male and 9998 female, for an overall gender ratio of 51.90\% male. 
However, within each class, the gender imbalance is much larger and the majority gender ratio taken as the weighted average across all classes is 61.54\%. 
In this situation, fairness issues may arise from the model associating the majority gender with certain actions. 

\subsection{Concept Bottleneck Model Setup}
\begin{figure*}[t]
    \centering
    \includegraphics[width=\textwidth]{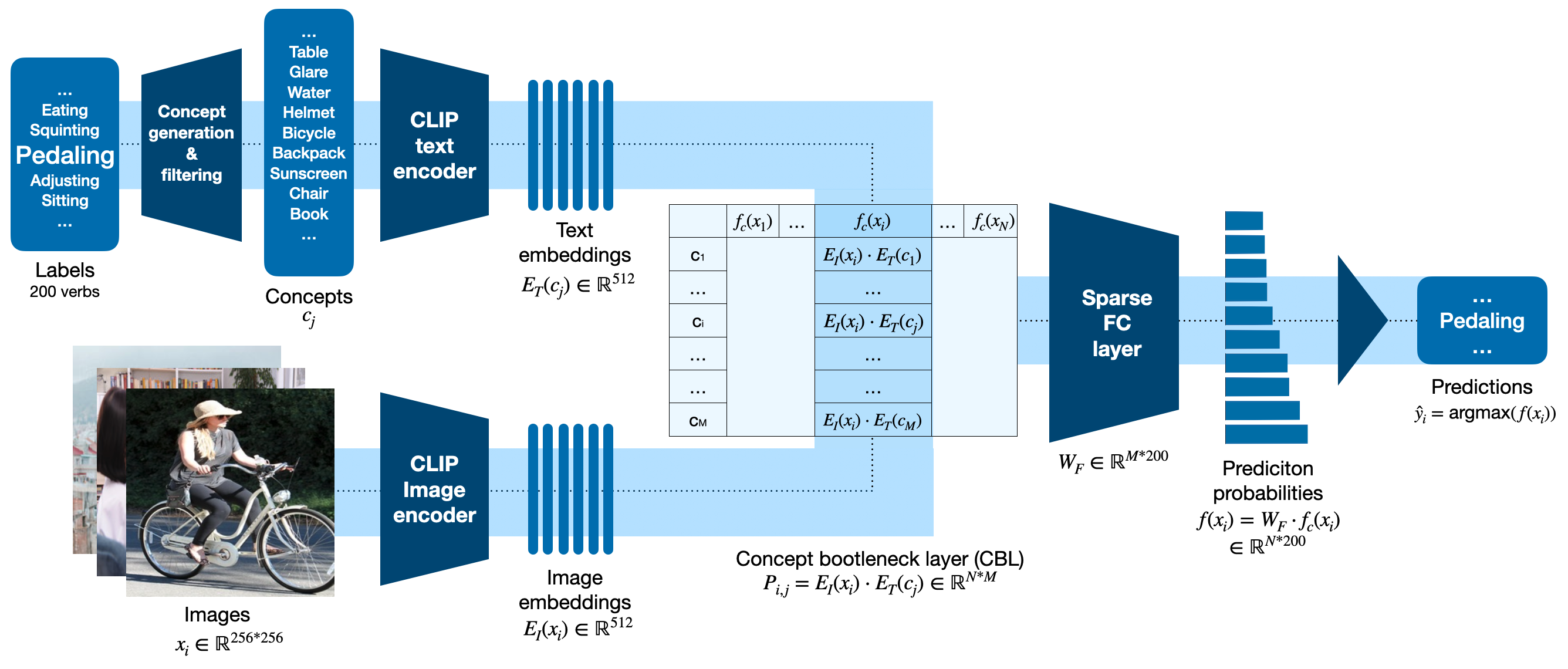}
    \caption{The architecture of our CBM with an image from ImSitu for the class pedaling.}
    \label{fig:CBM_arch}
\end{figure*}

We leverage a modified version of the label-free CBM framework as our CBM model. 
Although CBMs provide many advantages such as the ability to incorporate domain knowledge, interpretability, data efficiency, and robustness, the data requirements on concept generation and labeling make them harder to apply in practice. 
Label-free CBM alleviates the data requirement by generating concepts automatically using a large language model and using CLIP's zero-shot learning to perform concept inference. 
This allows the framework to generalize to a much wider range of image classification tasks as long as the concepts can be described. 
Figure \ref{fig:CBM_arch} illustrates our modified label-free CBM architecture.
We replaced the backbone neural network with CLIP's image encoder to increase efficiency and create a better comparison across different experiment setups. 

In the concept generation and filtering step, we queried the GPT-3 model with the 200 ImSitu classes (Verbs) using the following three prompts to generate a list of concepts: 1. List the most important features for recognizing something as (Verb), 2. List the things most commonly seen around (Verb), and 3. Give a superclass for the word (Verb). 
In order to increase the quality and decrease the length of the list of raw concepts, we removed concepts using the following four criteria sequentially. 
\begin{itemize}
    \item \textbf{Length} Concepts longer than 30 characters.
    \item  \textbf{Too similar to classes} Concepts with a cosine similarity $>$ 0.85 to any of the classes to prevent proxies.
    \item \textbf{Too similar to other concepts} Concepts with a cosine similarity $>$ 0.9 to another concept to ensure that each concept covers a unique space in the embedding.
    \item \textbf{Low activation on dataset} Concepts with top-5 highest image activations below an interpretability threshold do not apply to the dataset well.  
\end{itemize}

After finalizing the list of concepts, we encode the images using Google's vision transformer ViT-B/16 and the concepts using CLIP's tokenizer and text encoder. 
Both encoders are pre-trained on 400 million image-text pairs collected from the Internet to create a dual latent space. 
This step creates a matrix, known as the concept bottleneck layer, of size N images times M concepts in which each image is represented by its concept activations, a vector of length M. 
We then train a sparse, fully connected layer mapping the M concepts to the target label classes. 
Since sparsity facilitates interpretability by concentrating and highlighting activated concepts, we apply regularization on both the L1 and L2 norm of the weights~\cite{oikarinen2023label}, displayed in Equation \ref{CBM_loss}.
\begin{equation}
    \label{CBM_loss}
\min_{W_F, b_F} \sum_{i=1}^{N} L_{ce}(\hat{y_i}, y_i) + \lambda \left[ (1 - \alpha) \frac{1}{2} \|W_F\|_2^2 + \alpha \|W_F\|_1 \right]
\end{equation}
We denote the full CBM framework presented above as CLIP-CBM and introduce two other architectures for baseline comparisons.
CLIP-DNN treats CLIP's image encoder as a feature extractor and performs end-to-end deep learning by training a dense fully connected layer. 
This setting does not rely on concepts and represents the best possible performance by a neural network; there exists many choices for the pre-trained feature extractor, we used CLIP's encoder to achieve a better comparison with CLIP-CBM. 
The CLIP-ZS setting does not require any training and leverages CLIP's zero-shot learning to predict the 200 target classes directly. 
The setting allows us to measure CLIP's inherent bias versus the bias introduced during the training process. 

\subsection{Bias Mitigation Techniques}
Since CBMs are not designed to achieve fairness, we introduce three bias mitigation techniques on top of the proposed CBM framework: 1. Decreasing information leakage, 2. Removing biased concepts, and 3. Adversarial debiasing. 
These methods can be applied individually or in conjunction. 

Theoretically, CBMs improve fairness by replacing potentially bias inducing details with interpretable concepts. 
In practice, the concept activation vectors often encode information and hidden patterns beyond the concept semantics, known as information leakage. 
Label-free CBM reduces information leakage through sparsity, forcing the model to make strong associations between concepts and classes only. 
We propose a different approach using a top-k concept filter and quantization on the concept activation vectors. 
This resembles a human's mental model more closely - focusing on a much smaller set of prominent features and making estimations instead of calculating exact probability values. 
Decreasing information leakage does not rely on sensitive attribute labels for debiasing and can improve in multiple bias directions simultaneously. 

Although prompts to the large language model are neutral, generated concepts can contain proxy variables to the sensitive attributes. 
Manually examining the concept list and removing such instances can improve fairness. 
In practice, this process may be time consuming and will likely introduce additional human bias through removal decisions. 
A scalable and automated approach leverages the language model to self-rate the degree of association with sensitive attributes and remove concepts based on a cutoff threshold. 
Another approach involves training the CBM to predict gender and removing concepts with the highest weights. 

To optimize for best fairness-performance tradeoffs, we incorporate adversarial debiasing in the FC layer.
Doing so requires sensitive attribute labels in the given bias direction which may be unavailable in practice.
In this setup, CLIP-CBM is trained to perform the classification task whereas an adversary tries to predict the sensitive attribute from the model's output.  
The dual-objective optimization encourages strong model performance while remaining invariant to the sensitive attribute.
Compared to adversarial debiasing on a black-box DNN, this interpretable approach allows us to detect hidden dataset bias and collect new images more efficiently by examining how the concept weights shift. 

\section{Evaluation}

\subsection{Concept Generation and Training Details}
Generating concepts using GPT-3 on class names and performing the first three filtering steps produces 1976 concepts before setting the interpretability threshold. 
Table \ref{tab:exampl_concepts} provides examples of generated concepts, most of which resemble objects or locations.

Training details mostly revolve around the sparse FC layer. 
The weights $W_F$ were initialized using a Kaiming uniform initialization and the optimization problem was solved using a stochastic gradient descent (SGD) with a learning rate of $10^{-3}$ and batch size of 800.
The L2/L1 penalty ratio $\alpha$ was set to 0.99 following~\cite{oikarinen2023label}. 
The regularization parameter $\lambda$ and interpretability cutoff have been set to $10^{-3}$ and 0.25 respectively.

\begin{table}[t]
\centering
\caption{An example of 32 concepts generated from ImSitu's classes after filtering.}
\begin{tabular}{cccc}
\hline
Key & Dance & Cupboard & Backpack \\ 
Kettle & Perfume & Soccer ball & Shovel \\
Fan & Piano & Pair of gloves & Leash \\
Cat & Bathrobe & Pair of skates & Blanket \\
Beach & Clock & Music stand & Mirror \\
Laptop & Potted plant & Whiteboard & Headset \\
Scarf & Bar of soap & Makeup brush & Bookmark \\
Tree & Sculpture & Photo album & Music CD\\
\hline
\end{tabular}
\label{tab:exampl_concepts}
\end{table}

\subsection{Fairness Metrics}
We quantify fairness by leveraging the leakage and bias amplification metrics~\cite{wang2019balanced} for better comparison with prior work.  
The authors defined bias amplification as how much a model leaks information on the sensitive attribute when compared against another model of equal performance whose errors are only due to chance~\cite{wang2019balanced}. 
The metric captures instances when the model associates proxy features of sensitive attributes with the classes, a common cause for bias in image classification that cannot be mitigated even after artificially balancing the data. 
More specifically, dataset leakage measures how well one can predict gender without looking at the images and represents the bias inherently present in the dataset before training. 
Mathematically, this is very similar to the weighted average of the majority gender ratio of each class. 
Model leakage measures the bias after training the model. 
We can calculate this by replacing the ground truth class labels with predicted labels $\hat{y_i}$ and performing the same calculation as dataset leakage. 
Bias amplification is then defined as model leakage minus performance adjusted data leakage; this is achieved by randomly perturbing ground truth labels until reaching the same F1 score as the model. 
To account for the randomness in measuring bias amplification, we average results from 5 runs to increase precision. 
Compared to metrics such as differences in accuracy or false positive rates between gender, bias amplification provides a stronger estimation of systematic bias, focusing on what the model learns rather than how often it is correct.  

\subsection{Fairness-Performance Tradeoffs of CBM Models and Information Leakage}
To study whether CBMs improve fairness by masking unnecessary image details, we analyze the fairness-performance tradeoffs of CLIP-ZS, CLIP-DNN, and CLIP-CBM, shown in Table \ref{tab:CBM_perf}. 
As expected, CLIP-ZS does not take advantage of the training data and performs the worst in terms of accuracy. 
It also amplifies bias the least, although the actual value is quite far from zero, indicating that CLIP has a certain degree of inherent bias. 
CLIP-DNN achieves the highest performance by solving the optimization problem using a deep neural network but also amplifies bias the most during the process.  
In terms of accuracy, CLIP-CBM performs slightly worse than CLIP-DNN, indicating that CBMs can serve as an interpretable replacement for DNNs in tasks as long as there is a method for generating and labeling concepts. 
However, CLIP-CBM only slightly outperform CLIP-DNN in terms of bias amplification and is significantly worse than the inherent bias in CLIP.
Since the majority of concepts are not related to gender, these results suggest that information leakage may be occurring. 

\begin{figure}[t]
    \centering
    \includegraphics[width=0.98\textwidth]{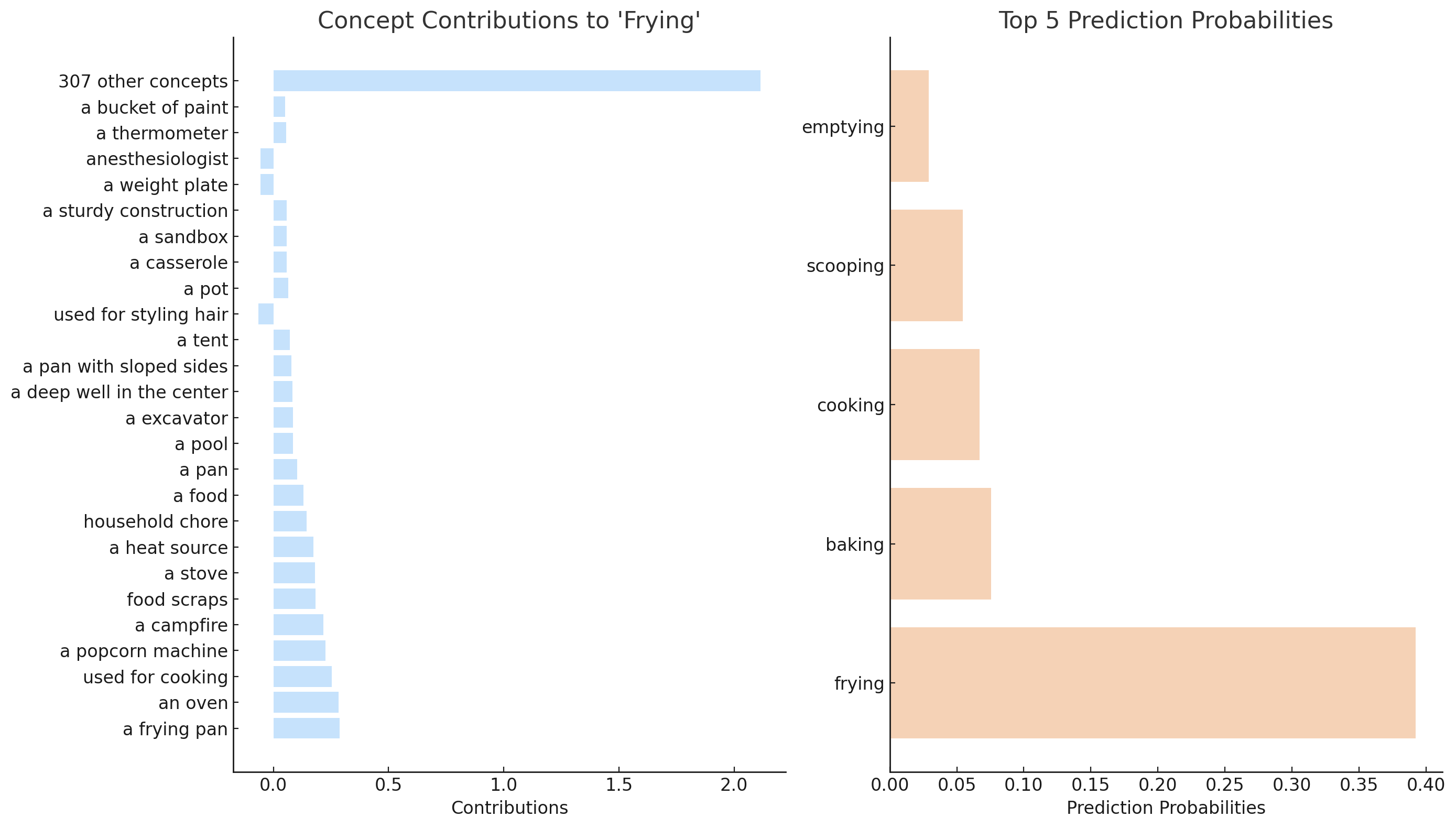 }
    \centering
    \includegraphics[width=0.98\textwidth]{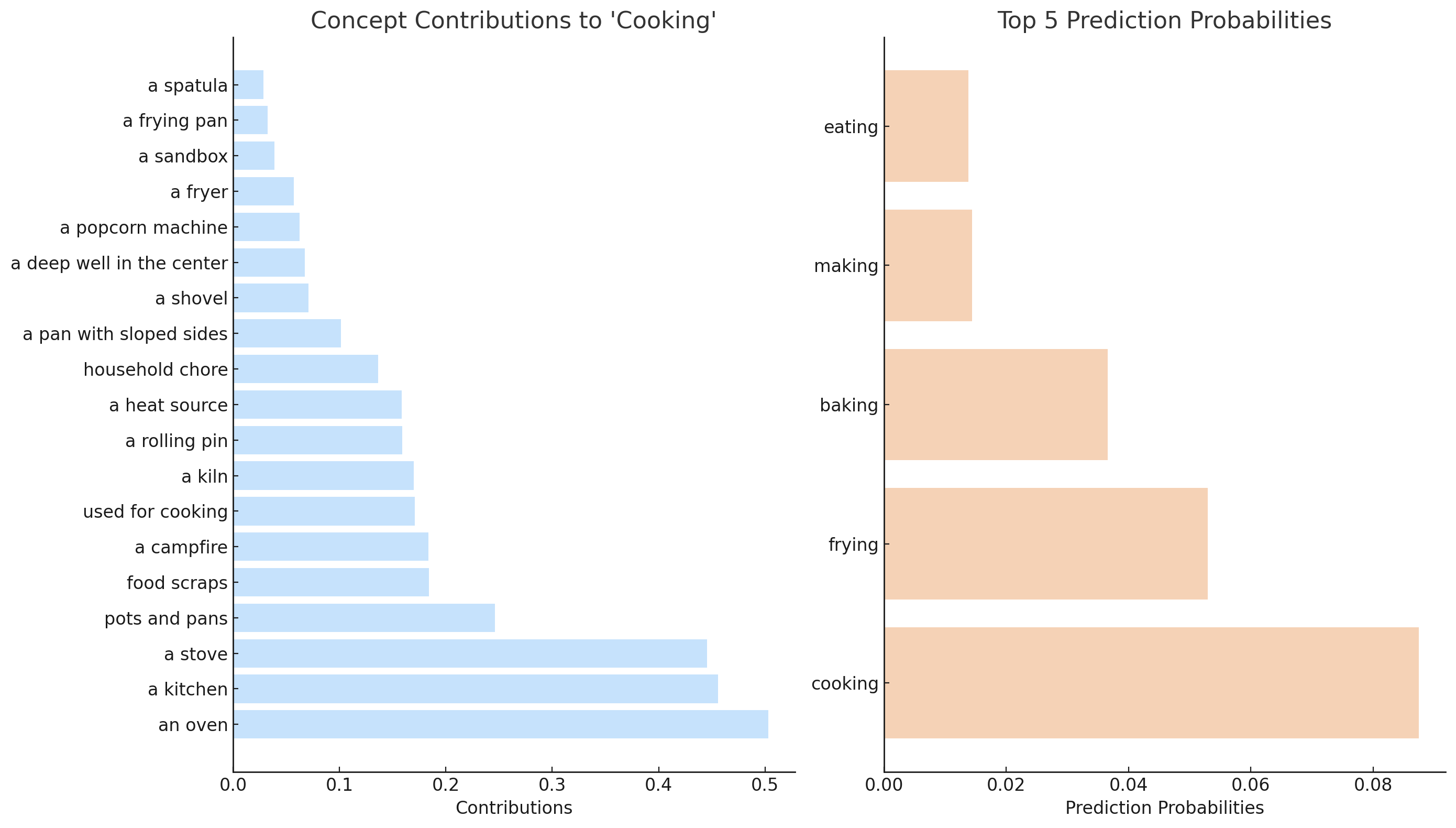 }
    \caption{Concept contributions and class predictions for an example image in 'frying' at different settings - using all concept contributions (top) and only the top 25 concept contributions each class (bottom) for prediction.}
    \label{fig:compare_sparse}
\end{figure}

We illustrate information leakage by describing how CBMs work in more detail. 
The sparse FC layer's concept weights provide model level information on how the concepts contribute to each class. 
Multiplying the concept weights by the corresponding concept activations yields concept contributions, which can provide image level information on important concepts that support the prediction.
Figure \ref{fig:compare_sparse} compares the concept contributions and predictions of an example image in the class 'frying' when we take into consideration all the concept contributions versus only the top 25 during prediction. 
We observe that the prediction is correct and the highest value concept contributions make sense semantically, focusing on 'a stove', 'a kitchen', and 'an oven'.
There are 332 non-zero concept contributions due to sparsity and the lower value concept contributions are numerous and harder to interpret, with the lowest 307 values summing up to similar magnitude as the top 8. 
However, if we make predictions based only on the top 25 concept contributions, the model incorrectly predicts the image as 'cooking' with the correct label 'frying' ranked second. 
This example illustrates that: 1. Interpreting only the top concept contributions suffers from a lack of faithfulness to model behavior and 2. Although many concepts are not related to 'frying' and do not contribute much to the prediction, together they encode a hidden distribution that can significantly affect the prediction.   

\begin{table}[t]
\centering
\caption{Comparisons of fairness-performance tradeoffs for CLIP-ZS, CLIP-DNN, and CLIP-CBM.}
\begin{tabular}{ccc}
\hline
Model & Accuracy & Bias Amplification \\
\midrule
CLIP-ZS & 30.74\% & 7.12\% \\ 
CLIP-DNN & 44.10\% & 8.68\% \\ 
CLIP-CBM & 41.51\% & 8.19\% \\ 
\hline
\end{tabular}
\label{tab:CBM_perf}
\end{table}

\subsection{Improvements after Bias Mitigation}
\subsubsection{Technique 1: Decreasing Information Leakage}
\begin{figure}[t]
    \centering
    \includegraphics[width=0.8\linewidth]{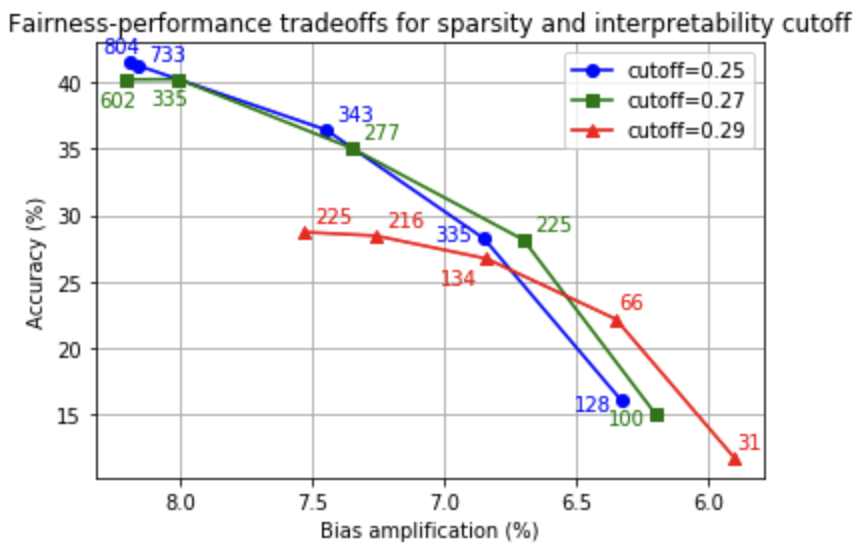}
    \caption{Fairness-performance tradeoffs of models with different $\lambda$ (0.05, 0.01, 0.005, 0.001, and 0.0005) and interpretability threshold cutoffs (0.25, 0.27, and 0.29) with the number of non-zero concept weights averaged across classes included.}
    \label{fig:sparsity_pareto}
\end{figure}

Decreasing information leakage can improve both fairness and interpretability. 
To study how this affects performance, we trained 15 models using different combinations of the sparsity regularization parameter $\lambda$ and interpretability threshold for concept filtering, displayed in Figure \ref{fig:sparsity_pareto}.
In general, there exists a tradeoff between fairness and performance with more sparse models increasing fairness at the expense of accuracy and vice versa. 
We observed a very strong correlation between the number of non-zero concept contributions and performance, ranging from close to an optimized DNN at 804 average concepts to around 12\% at 31 concepts. 
This indicates that the large number of concept contributions behind information leakage also play an important role in accurately capturing the concept-class mapping by providing a certain degree of complexity. 
Decreasing the number of concepts using the interpretability threshold improves parameter efficiency (better fairness-performance tradeoffs with less concepts) but reaches a performance limit. 

\begin{figure}[t]
    \centering
    \includegraphics[width=0.8\linewidth]{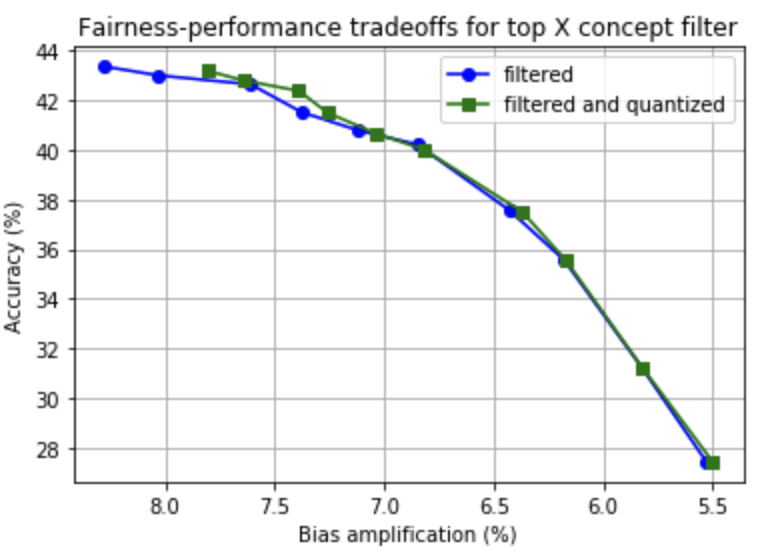}
    \caption{Fairness-performance tradeoffs of models with a top-k concept activation filter, with k values: 5, 10, 20, 30, 50, 70, 100, 200, 500, 1000.}
    \label{fig:cutoff_pareto}
\end{figure}

To improve on these results, we propose an alternative to sparsity by using a top-k concept activation filter; Figure \ref{fig:cutoff_pareto} displays fairness-performance tradeoffs for different values of k.
Instead of decreasing the number of concept weights in the FC layer, we retain the values of the top k concept activations only and zero out the remainder in the concept activation vectors; the number of concept contributions for each image equals k. 
Results still indicate a tradeoff between fairness and performance with a larger k value having better accuracy. 
However, the optimal curve of fairness-performance tradeoffs has improved considerably.
At k = 1000, the model performance even approaches that of CLIP-DNN while having lower bias amplification.
In terms of parameter efficiency, k = 30 achieves an accuracy of 37.5\% which is significantly higher than the 12\% from 31 average concept contributions in Figure \ref{fig:sparsity_pareto}.
We also applied a simplistic quantization algorithm to the concept activation vector, grouping values every 0.5 standard deviations above average activation into the same bucket. 
Doing so decreases the model's ability to learn the hidden distribution from lower valued concept contributions to prevent information leakage. 
Although results are insignificant when k is low, quantization provided some improvement to fairness at minimal accuracy loss when the number of concept contributions increases. 

These findings suggest that the top-k concept activation filter greatly outperforms sparsity in terms of bias mitigation. 
Nevertheless, there exists a fundamental tradeoff between fairness, interpretability, and performance. 
A CBM with strong performance requires a certain number of concept contributions yet these concepts will leak information on the sensitive attribute beyond concept semantics and make it harder to interpret results. 

\subsubsection{Technique 2: Removing Biased Concepts}
Removing concepts that serve as proxy variables to sensitive attributes can improve fairness. 
To illustrate this, we trained the CBM to predict gender instead of the ImSitu classes and sorted the concept weights in descending order to identify the most biased concepts. 
For male, the top 10 concepts are: ['a chuck', 'a jerk', 'clay', 'a mat', 'a football', 'cloak and dagger type clothing', 'usually done with force', 'a surgeon', 'a necktie', 'a tie rack'] and for female: ['a nurse', 'a dolly', 'a midwife', 'a pointed top', 'a blouse', 'a receptionist desk', 'a blow dryer', 'a goal', 'a mixer', 'a therapist']. 
Concepts such as 'necktie', 'tie rack', 'pointed top', and 'blouse' serve as visual gender proxies through clothing and should be removed.
However, we also found regular objects such as 'clay', 'mat', 'goal', and 'dolly' as well as concepts that may improve fairness at the cost of performance for certain classes, such as 'nurse', 'football', 'receptionist desk', and 'blow dryer'. 
Removal decisions for these concepts are more complicated and should include factors such as information leakage, fairness-performance tradeoffs, feature correlations pertaining to the specific dataset, etc. 
A simpler approach that does not require model training and sensitive attribute ground truth focuses only on the concept semantics. 
Since manually examining all 1976 concepts requires intensive resources, we leveraged the LLM's self ratings and removed the 46 most biased concepts. 

After removing the list of biased concepts and re-training the CBM, we are surprised that the model did not improve in fairness. 
We hypothesize that this is due to information leakage: The new model learns a new hidden distribution from the concept activation vectors and its weights leak gender information in a different way. 
This phenomena implies that concept semantics play a weaker role in fairness than previously expected as all concepts leak information on the sensitive attribute albeit to different degrees. 
To overcome this situation, we zeroed out the concept activations of removed concepts at testing time and performed inference using the old model instead. 
Results indicate that removing the top 50 biased concepts from the gender predictor and the top 46 biased concepts based on LLM self ratings decrease bias amplification by 0.5\% and 0.3\% respectively while incurring around 0.6\% loss in accuracy. 
In general, this method does not achieve strong tradeoffs and its justifications are challenged by information leakage. 

\subsubsection{Technique 3: Adversarial Debiasing}
Adversarial debiasing optimizes for fairness-performance tradeoffs by minimizing the leakage of gender information across the concept-class mapping while preserving performance.
Compared to removing biased concepts for the entire dataset, adversarial debiasing provides more granular improvements, affecting certain classes only. 
Table \ref{tab:adv_perf} displays fairness-performance tradeoffs for CLIP-DNN and both the sparsity and top-k concept filter versions of CLIP-CBM before and after adversarial debiasing. 
The debiased CLIP-DNN provides comparison with prior work~\cite{wang2019balanced} while removing the effects of different pre-trained feature extractor networks; adversarial debiasing was instead performed on the CLIP embedding-class mapping. 
For all three models, adversarial debiasing incurred a loss in accuracy between 0.7\% and 1.3\%. 
In terms of fairness, all models improved around 1.5\%; note that the improved bias amplification score is close to or better than that of the pre-trained CLIP zero-shot model.
These results show that 1. The fairness improvements from adversarial debiasing can be added to those gained from adopting a lower information leakage CBM and 2. The fairness-performance tradeoffs of adversarial debiasing are better than removing concepts. 

\begin{figure}[t]
    \centering
    \includegraphics[width=0.8\linewidth]{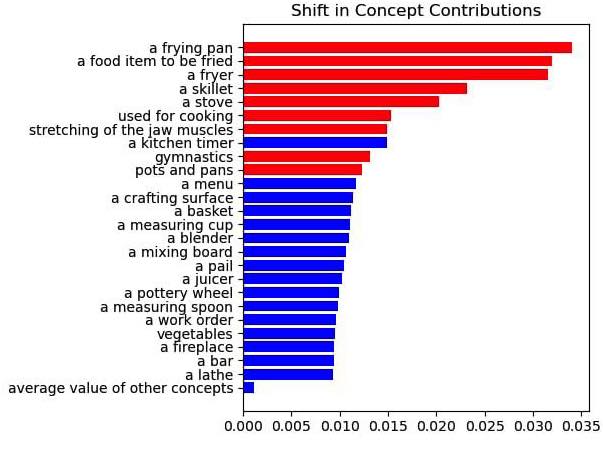}
    \caption{Shifts in class averaged concept contributions for 'frying' before and after applying adversarial debiasing to CLIP-CBM. Values sorted in descending order by magnitude, red indicates increases (blue decreases) after debiasing.}
    \label{fig:adv_shifts}
\end{figure}

Compared to applying adversarial debiasing on a black-box DNN, the debiased CBM benefits from image level interpretability, model level interpretability, and an interpretable debiasing process. 
Making the debiasing process transparent provides many benefits, including increasing trust in model behavior, preventing fairness gerrymandering~\cite{kearns2018preventing}, and stopping bias from increasing in other directions. 
In the ideal case, we should observe weight decreases in concepts that help support the class prediction but are more strongly associated with one gender. 
In a hypothetical example, bias occurs when for the class 'painting', male images contain more samples of wall painting and female images more samples of art painting. 
Debiasing would then decrease the weights of 'ladder', 'roof', and 'artist' while increasing weights for 'brush' and 'paint'.
In our experiments, we noticed that some of the high magnitude concept weight shifts are unrelated to 'frying', reinforcing earlier findings that CBM concepts leak information beyond word semantics. 
In Figure \ref{fig:adv_shifts}, we calculate the changes in class averaged concept contributions caused by adversarial debiasing for the class 'frying'. 
To understand behavior at the model level, we compute the average of concept contributions for all images predicted to belong to the class. 
We see highest increases in concepts clearly associated with 'frying' and high decreases for concepts that are related to 'frying' but may also support alternative decisions, i.e. 'a kitchen timer' may be used for 'cooking' as well. 
This suggests that adversarial debiasing achieves strong fairness-performance tradeoffs by focusing on meaningful concepts and decreasing the confusion matrix's gender correlation. 

\begin{table}[t]
\centering
\caption{Fairness-performance tradeoffs for models before and after applying adversarial debiasing.}
\begin{tabular}{ccc}
\hline
Model & Accuracy & Bias Ampl. \\
\midrule
CLIP-DNN & 44.10\% & 8.68\% \\ 
CLIP-DNN + debiasing~\cite{wang2019balanced} & 42.83\% & 7.21\% \\
CBM-sparse & 41.51\% & 8.19\% \\ 
CBM-sparse + debiasing & 40.06\% & 6.87\% \\ 
CBM-topk & 43.35\% & 7.84\% \\ 
\textbf{CBM-topk + debiasing} & \textbf{42.69\%} & \textbf{6.29\%} \\
\hline
\end{tabular}
\label{tab:adv_perf}
\end{table}

\section{Discussion}
Although this paper focuses on fairness and bias mitigation, the findings provide valuable insights to users of CBMs and the study of interpretability.
By providing a clear problem setting, including a real-world dataset, a task, quantitative metrics, and human understandable contexts and concepts, gender bias can be used to evaluate the efficacy and promises of CBMs. 
Our results show that CBM concepts leak information beyond concept semantics, creating a tradeoff between fairness, interpretability, and performance. 
Analyzing information leakage illustrates the inner workings of CBM models. 
The concept activation vector can be viewed as a point in a high dimensional embedding space where each dimension provides more interpretability than a regular neural network by loosely resembling human understandable concepts.
The CBM essentially performs a transformation from the CLIP embedding space to the concept embedding space. 
The main benefit of the approach lies in the built-in interpretability between the concept-class mapping, eliminating the errors introduced by post-hoc explanation mechanism's approximations. 
However, neither the concept representations nor the embedding space transformation are perfect so the CBM remains not perfectly transparent. 

Throughout this paper, we used CBM and CLIP-CBM, a modified version of the label-free CBM framework, interchangeably.
For most real-world problems, traditional CBMs have very limited scalability due to efforts required for concept generation and labeling, an area where label-free CBM overcomes using GPT and CLIP. 
However, our findings on the tradeoffs described in Section 4.4 suggest that label-free CBMs may be less interpretable than their traditional counterparts by design. 
Since GPT generates concepts using general prompts, the precision and descriptiveness are worse than those of a human expert.
We also noticed that some concepts are less visual and more abstract, rewarding semantical similarity instead of visual evidence; for example 'arms embraced' will likely serve as a better alternative to 'love'. 
As a result, label-free CBMs require more concepts to achieve the same performance, making interpretation of concept weights and contributions more difficult and increasing information leakage. 
Furthermore, CLIP's zero-shot learning for concept activations are imperfect and captures existing societal bias. 
Rather than producing a boolean value or ordinal levels, the less-confident scores ranging around 0.1 to 0.25 create lengthy and noisy concept activation vectors.
In general, improving CBMs remains an active field of research and fairness can serve as a dimension for evaluating new approaches. 

Despite some of the aforementioned CBM limitations, they still provide valuable contributions to the field of fairness. 
As long as one can limit information leakage, for example by using the top-k concept filter version of CLIP-CBM that we proposed, CBMs improve both fairness and interpretability while only incurring minimal accuracy loss. 
This approach of fairness through unawareness does not rely on sensitive attribute ground truth labels, which may unavailable, error-prone, or expensive to collect in practice. 
The same benefit applies to scenarios when multiple sensitive attributes should be considered but some are unobserved, for instance race and age for the ImSitu dataset. 
Since most bias mitigation algorithms target one sensitive attribute, CBMs are more likely to improve or at least not exacerbate bias in other directions. 
To reduce bias further, one can remove biased concepts based on semantics. 
If the sensitive attribute is clear and labeled for input images, then applying adversarial debiasing to CBMs optimize for fairness-performance tradeoffs; we do not recommend removing concepts in this case.
Doing so can also provide insights on the cause of bias by examining changes in the concept weights and contributions during debiasing. 

\section{Conclusion}
In this paper, we mitigate bias in CBMs to enable fair and interpretable image classification. 
We evaluated results on ImSitu, a real-world action recognition dataset, and implemented a modified version of the label-free CBM framework, which overcomes challenges in concept generation and labeling through GPT and CLIP. 
Since CBMs are not designed to achieve fairness, we propose three bias mitigation techniques for CBM frameworks: 1. Decreasing information leakage, 2. Removing biased concepts, and 3. Adversarial debiasing.

We found that CBMs improve both fairness and interpretability compared to black-box DNNs, at the expense of a slight decrease in performance. 
However, CBM concepts also encode hidden information unrelated to concept semantics. 
In general, CBMs do not achieve perfect transparency because they essentially perform a transformation into the high dimensional concept embedding space; this issue is more prevalent in label-free CBM than traditional CBMs. 
This creates a tradeoff between fairness, interpretability, and performance: The model requires a certain number of concepts to achieve strong performance but doing so leaks information on the sensitive attributes and makes interpreting concept contributions more difficult. 
The top-k concept filter we proposed provides better fairness-performance tradeoffs than applying sparsity to the FC layer but still does not solve information leakage entirely.  
Removing biased concepts, either using a gender classifier or based on semantics alone, do not produce strong results and are limited by information leakage. 
Adversarial debiasing optimizes for fairness-performance tradeoffs, gaining further fairness improvements on top of CBM models. 
Our best model, top-k concept filter with adversarial debiasing, decreases bias amplification by 28\% with very minor accuracy loss. 
Examining the shifts in concept weights and contributions also provides more interpretability and transparency in the debiasing process. 
Overall, our bias mitigation techniques substantially enhance fairness-performance tradeoffs beyond prior approaches, marking a decisive advancement toward fairer and more interpretable image classification.


\bibliographystyle{ACM-Reference-Format}
\bibliography{sample-base}


\end{document}